\documentclass{article}

\usepackage[utf8]{inputenc} 
\usepackage[T1]{fontenc} 
\usepackage[english]{babel} 

\usepackage{amsmath} 
\usepackage{amssymb} 
\usepackage{amsthm} 
\usepackage{mathtools} 
\usepackage{bm} 

\usepackage{graphicx} 
\usepackage{float} 
\usepackage{caption} 
\usepackage{subcaption} 
\usepackage{wrapfig} 
\usepackage{epstopdf} 

\usepackage{booktabs} 
\usepackage{array} 
\usepackage{tabularx} 
\usepackage{multirow} 
\usepackage{longtable} 
\usepackage{ltablex} 

\usepackage{hyperref} 
\usepackage{cleveref} 
\usepackage[numbers]{natbib} 

\usepackage{comment} 
\usepackage{todonotes} 

\usepackage[bottom]{footmisc} 

\usepackage{geometry} 
\usepackage{enumitem} 
\usepackage{siunitx} 
\usepackage{url} 
\usepackage{fancyhdr} 
\usepackage{setspace} 
\usepackage{titlesec} 
\usepackage{xcolor} 

\title{Semi-Supervised Fine-Tuning of Vision Foundation Models with Content-Style Decomposition}
\author{
    \makebox[\textwidth][c]{ 
        \begin{tabular}{ccc}
            \textbf{Mariia Drozdova} & \textbf{Vitaliy Kinakh} & \textbf{Yury Belousov} \\
            \texttt{mariia.drozdova@unige.ch} & \texttt{vitaliy.kinakh@unige.ch} & \texttt{yury.belousov@unige.ch} \\
        \end{tabular}
    } \\
    \makebox[\textwidth][c]{
        \begin{tabular}{cc}
            \textbf{Erica Lastufka} &  \textbf{Slava Voloshynovskiy\footnote{Slava Voloshynovskiy (svolos@unige.ch) is the corresponding author.}} \\
            \texttt{erica.lastufka@unige.ch} &  \texttt{svolos@unige.ch} \\
        \end{tabular}
    } \\
    \textit{University of Geneva, Switzerland}
}
\date{}

\begin{document}

\maketitle

\begin{abstract}  In this paper, we present a semi-supervised fine-tuning approach designed to improve the performance of pre-trained foundation models on downstream tasks with limited labeled data. By leveraging content-style decomposition within an information-theoretic framework, our method enhances the latent representations of pre-trained vision foundation models, aligning them more effectively with specific task objectives and addressing the problem of distribution shift. We evaluate our approach on multiple datasets, including MNIST, its augmented variations (with yellow and white stripes), CIFAR-10, SVHN, and GalaxyMNIST. The experiments show improvements over supervised finetuning baseline of pre-trained models, particularly in low-labeled data regimes, across both frozen and trainable backbones for the majority of the tested datasets. \end{abstract}

\section{Introduction}

In recent years, vision foundation models have gained significant popularity across a wide range of applications. These models, which are typically pre-trained on vast publicly available datasets, have demonstrated their utility in numerous domains, including multimedia applications and scientific fields such as astronomy\cite{lastufka2024selfsupervisedlearningmeerkatwidefield}, biology\cite{stevens2024bioclip}, medical imaging\cite{zhang2023challenges}, and remote sensing\cite{lu2024ai}. The common practice in developing these vision foundation models involves pre-training them in a self-supervised (DINOv2 \cite{oquab2023Dinov2}) or weakly supervised (CLIP \cite{ilharco_gabriel_2021_5143773, Radford2021LearningTV}, RADIOv2\cite{Ranzinger_2024_CVPR}) manner. These training methodologies do not target any specific downstream tasks; instead, they aim to find a representation that can be adapted to a variety of tasks. This characteristic supports the versatility of foundation models but introduces challenges when applying them to downstream tasks (DSTs), particularly when there is a distribution mismatch between the pre-training data and the DST data.

\subsection{Challenges in Fine-Tuning Foundation Models for Downstream Tasks}

One of the primary challenges in utilizing vision foundation models for specific DSTs is the issue of {\bf distribution shift}. The distribution of data $p_{\mathbf{x}}^{\operatorname{tr}}(\mathbf{x})$ used during the pre-training phase of vision foundation models does not necessarily match the distribution of the data $p_{\mathbf{x}}(\mathbf{x})$ associated with a particular DST. This mismatch can lead to suboptimal performance when the foundation model is directly applied to DSTs without appropriate adjustments. {Our proposed semi-supervised learning approach aims not only to adapt the foundation model to the task objective but also to address this distribution shift by better aligning the learned representations with DST data distributions.} Consequently, fine-tuning becomes essential to adapt the model's learned representations to align with the characteristics of the DST data.

Furthermore, the pre-training of vision foundation models often aligns with the {\bf principles of the information bottleneck theory}\cite{tishby2000information}, which suggests that a model should retain only the information relevant to a particular task while discarding irrelevant details. However, self- or weakly-supervised learning, by its nature, might either retain too much or too little information relevant to a specific DST since DST is unknown at the pre-training stage. When faced with distribution mismatches between DST and pre-training, the model may struggle to retain the appropriate information for downstream tasks. In cases where essential information is overly suppressed, fine-tuning might not fully recover the necessary details. Conversely, if too much information is retained, targeted fine-tuning can help prune and specialize the model's representations for the DST.

\subsection{Fine-Tuning Challenges in Scientific Applications}

In this study, we focus on fine-tuning vision foundation models for simple datasets, simulating scenarios often found in scientific applications where labeled data is limited. In these situations, we typically encounter a moderate number of unlabeled examples, ranging from 10,000 to 70,000, while the number of labeled examples may vary from just 1-2 to all available samples per class. Our approach demonstrates how leveraging both labeled and unlabeled data can help address these issues by improving the representations of foundation models in downstream tasks, even when distribution shift is present.

\section{Information-Theoretic Framework for Fine-Tuning of VFMs}

As an alternative to conventional fine-tuning approaches, we propose leveraging methods grounded in semi-supervised learning, a well-established field that has demonstrated its effectiveness in various contexts. Specifically, we extend an information-theoretic framework to address the fine-tuning challenge \cite{voloshynovskiy2020variational}. This framework enables the model to adapt to the specifics of the downstream task by optimizing the retention of task-relevant information while discarding irrelevant details.

Rather than applying this approach directly to image data, we focus on the latent representations of vision foundation models. By improving these representations, our method aims to overcome the distribution shift that occurs when applying models to downstream datasets by leveraging available unlabeled data and a limited number of labeled examples.

We demonstrate the applicability of this approach by evaluating its performance across datasets that are simpler and distinct from large-scale, high-resolution natural image distributions, such as MNIST dataset variations\cite{lecun1998mnist}, SVHN\cite{netzer2011reading}, CIFAR-10\cite{krizhevsky2014cifar}, and GalaxyMNIST\cite{walmsley2022galaxy}. More details about these datasets can be found in Section \ref{sec:datasets}. Our results indicate that while certain vision foundation models struggle with these simpler datasets—particularly in low-labeled sample cases with purely supervised fine-tuning approaches—others are better suited to these specialized applications. This highlights the importance of model selection and fine-tuning strategies. This work showcases the advantages of our proposed approach in enhancing the adaptability and effectiveness of foundation models for handling diverse data.

\subsection{Proposed Semi-Supervised Fine-Tuning Scheme}

\begin{figure}[h]
    \centering
    \includegraphics[width=0.9\textwidth]{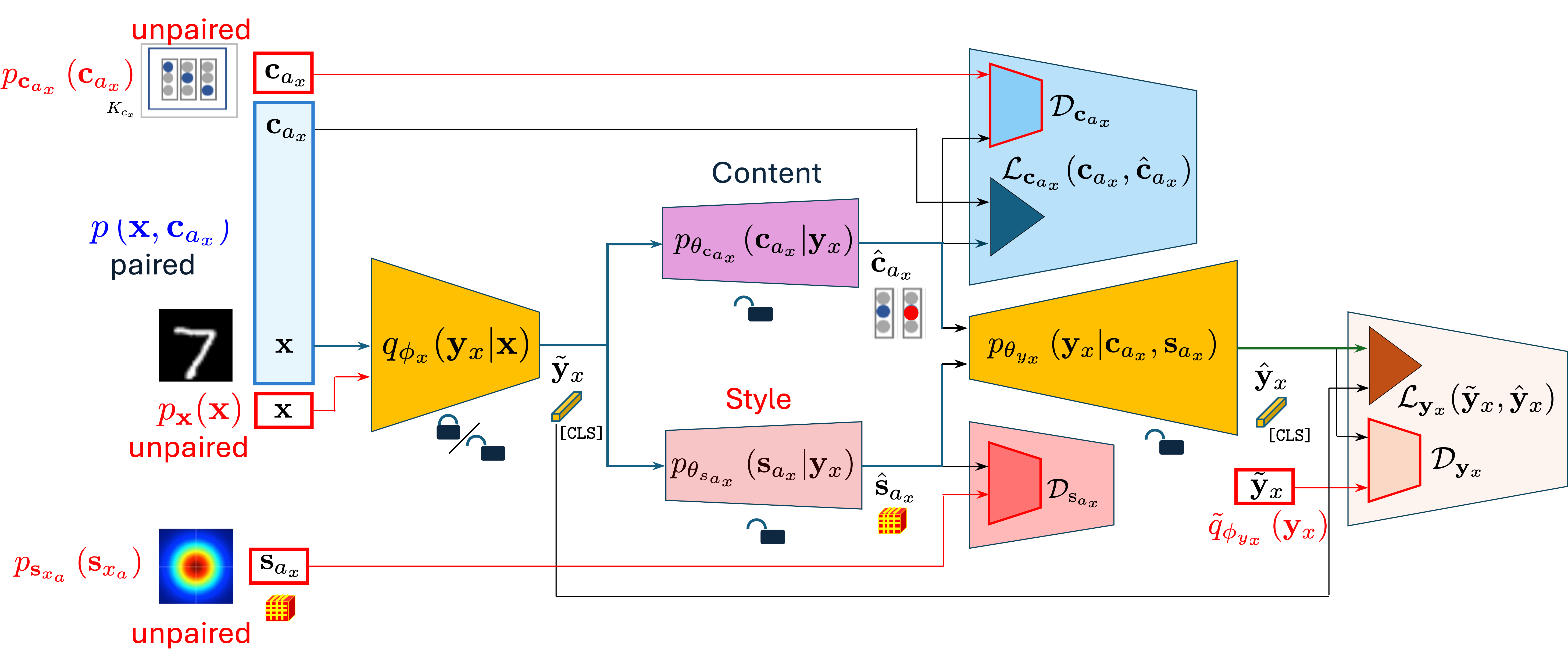}
    \caption{Architecture of the proposed semi-supervised fine-tuning scheme. The vision foundation model generates a representation in the form of a \texttt{[CLS]} token $\tilde{\mathbf{y}}_x$, which is decomposed into content attribute label $\hat{\mathbf{c}}_{a_x}$ and generic style $\hat{\mathbf{s}}_{a_x}$. These are then used for targeted reconstruction of the \texttt{[CLS]} token $\hat{\mathbf{y}}_x$.}
    \label{fig:model_architecture}
\end{figure}

The proposed semi-supervised fine-tuning method for the vision foundation model leverages the principle of content-style decomposition. The foundation model, denoted as an encoder $q_{\phi_x}\left(\mathbf{y}_x | \mathbf{x}\right)$, pre-trained on a large dataset with the distribution $p_{\mathbf{x}}^{\operatorname{tr}}(\mathbf{x})$, maps an input $\mathbf{x}$ to a latent representation $\tilde{\mathbf{y}}_{x}$. The latent representation can be in the form of patch tokens and a \texttt{[CLS]} token for transformer-based models, aggregation or summary tokens, or just a vector output of CNN-based foundation models. Since in this work DST task is a classification, we proceed with the \texttt{[CLS]} token representation. One might use patch tokens or CNN latent tensors for more complex tasks such as segmentation or depth estimation.  Therefore, the latent representation $\tilde{\mathbf{y}}_{x}$ in the form of a \texttt{[CLS]} token is then decomposed into content $\mathbf{c}_{a_x}$ and style $\mathbf{s}_{a_x}$ representations. As this work focuses on classification, the content $\mathbf{c}_{a_x}$ is represented as a one-hot encoding of the class. The content and style are then subsequently combined and passed through a decoder to reconstruct the \texttt{[CLS]} token $\hat{\mathbf{y}}_x$.

Each element of the model, namely content representation, style representation, and \texttt{CLS} token reconstruction, is associated with specific blocks and regularizers. We employ the concept of {\bf adversarial mutual information decomposition} \cite{voloshynovskiy2020variational}, which leads to the formulation where mutual information is decomposed into: (a) a conditional cross-entropy term, which serves as a metric of similarity between content, style, and the \texttt{CLS} token for the paired data, and (b) a discriminator term, which represents the Kullback-Leibler divergence for unpaired data. 

In the proposed semi-supervised model, the vision foundation model that generates the initial representations can be kept either \textbf{frozen} or \textbf{trainable}, depending on the setup.

\begin{itemize}
    \item In the \textbf{frozen setup}, only the content prediction block \(p_{\theta_{{c}_{a_x}}}\left(\mathbf{c}_{a_x} | \mathbf{y}_x\right)\), style prediction block \(p_{\theta_{{s}_{a_x}}}\left(\mathbf{s}_{a_x} | \mathbf{y}_x\right)\), and \texttt{CLS} token reconstruction \(p_{\theta_{y_x}}(\mathbf{y}_x|\mathbf{c}_{a_x}, \mathbf{s}_{a_x})\) components are trained, while the foundation model remains unchanged.
    \item In the \textbf{trainable setup}, both the foundation model \(q_{\phi_x}(\mathbf{y}_x|\mathbf{x})\) and the components \(p_{\theta_{{c}_{a_x}}}\left(\mathbf{c}_{a_x} | \mathbf{y}_x\right)\), \(p_{\theta_{{s}_{a_x}}}\left(\mathbf{s}_{a_x} | \mathbf{y}_x\right)\) and \(p_{\theta_{y_x}}(\mathbf{y}_x|\mathbf{c}_{a_x}, \mathbf{s}_{a_x})\) are trained, but with different learning rates: the backbone learning rate is set to be 100 times smaller than that of the classifier and other components to avoid excessive disruption of the foundation model’s pre-trained features. 
\end{itemize}

The core idea of the fine-tuning process in this semi-supervised model revolves around the availability of a certain number of paired examples. Let $\mathcal{X}$ represent the input data, and $\mathcal{C}_A$ denote the content attributes. We define $N_{\mathcal{X}, \mathcal{C}}$ as the total number of paired examples $\{\mathbf{x}_i, \mathbf{c}_{{a_x}_i}\}_{i=1}^{N_{\mathcal{X}, \mathcal{C}}}$, which is assumed to be limited. Additionally, let $N_{\mathcal{X}}$ denote the number of unpaired samples $\mathbf{x}$, derived from the downstream task distribution $p_{\bf x}({\bf x})$. Here, the downstream task is focused on content attribute prediction. 

The main mechanism of our system is as follows:

\begin{itemize}
    \item The content predictor, or content attribute estimator, is trained to predict the correct content attribute $\mathbf{c}_{a_x}$ for paired data $\{\mathbf{x}_i, \mathbf{c}_{{a_x}_i}\}_{i=1}^{N_{\mathcal{X}, \mathcal{C}}}$. This attribute is represented as a one-hot encoded class label.
    \item For unpaired data $\{\mathbf{x}_i\}_{i=1}^{N_{\mathcal{X}}}$, the system ensures that the predicted content attribute belongs to one of the predefined classes. The distribution of these attributes is modeled as a categorical distribution, regulated by the discriminator and its corresponding categorical representation of content attributes.
\end{itemize}

In this model, the style distribution is not explicitly constrained and is assumed to follow a Gaussian distribution. This approach allows the model to adapt flexibly to different styles while maintaining robustness in content attribute prediction.

As shown in Figure \ref{fig:model_architecture}, for paired data $\{\mathbf{x}_i, \mathbf{c}_{{a_x}_i}\}_{i=1}^{N_{\mathcal{X}, \mathcal{C}}}$, we apply a cross-entropy loss $\mathcal{L}_{c_{a_x}} \left( \tilde{\mathbf{c}}_{a_x}, \hat{\mathbf{c}}_{a_x} \right)$, which represents the supervised learning step. In contrast, for unpaired data $\{\mathbf{x}_i\}_{i=1}^{N_{\mathcal{X}}}$ the model performs content-style disentanglement and reconstruction, regulated by discriminators operating on different spaces. These discriminators, \(\mathcal{D}_{\mathbf{c}_{a_x}}\) for content, \(\mathcal{D}_{\mathbf{s}_{a_x}}\) for style, and \(\mathcal{D}_{\mathbf{y}_{a_x}}\) for the reconstructed \texttt{CLS} token, enforce Kullback–Leibler divergences on the corresponding spaces. Additionally, a reconstruction loss for the \texttt{CLS} token, denoted as $\mathcal{L}_{y_{a_x}} \left( \tilde{\mathbf{y}}_{a_x}, \hat{\mathbf{y}}_{a_x} \right)$, is calculated using cosine similarity between the true \texttt{CLS} token $\tilde{\mathbf{y}}_{a_x}$ and the predicted one $\hat{\mathbf{y}}_{a_x}$. These losses are integrated into an unsupervised learning step. Further architectural details and mathematical definitions can be found in Appendices \ref{appendix_A},  \ref{appendix_B}. 

This formulation follows the approach proposed in \cite{voloshynovskiy2020variational}, where detailed ablation studies were conducted. Based on these results, we selected the best setup for our work: learnable priors and enabling all proposed losses.

In the trainable setup, the backbone is updated only during the supervised step and not during the unsupervised steps. This is because we reconstruct the \texttt{CLS} token, and allowing gradients through both components and the backbone would lead to degenerate results. The learning process involves alternating between supervised and unsupervised updates: for the first 20 iterations, we use only supervised losses. Afterward, we introduce unsupervised losses, performing 1 unsupervised update for every 2 supervised updates. This configuration was selected after a brief hyperparameter search, where it showed optimal balance between leveraging labeled data and making effective use of the unlabeled data.

This semi-supervised fine-tuning scheme provides a robust framework for leveraging the expressive power of pre-trained vision foundation models while efficiently utilizing limited labeled data and a larger pool of unlabeled data to adapt to specific downstream tasks.





\subsection{Datasets}
\label{sec:datasets}
\begin{figure}[h!]
    \centering
    \setlength{\tabcolsep}{2pt} 
    \renewcommand{\arraystretch}{1} 
    \begin{tabular}{ccccccc} 
        \includegraphics[width=.13\textwidth]{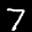} &
        \includegraphics[width=.13\textwidth]{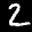} &
        \includegraphics[width=.13\textwidth]{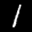} &
        \includegraphics[width=.13\textwidth]{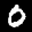} &
        \includegraphics[width=.13\textwidth]{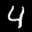} &
        \includegraphics[width=.13\textwidth]{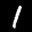} &
        \includegraphics[width=.13\textwidth]{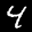} \\
        
        \includegraphics[width=.13\textwidth]{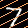} &
        \includegraphics[width=.13\textwidth]{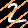} &
        \includegraphics[width=.13\textwidth]{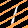} &
        \includegraphics[width=.13\textwidth]{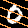} &
        \includegraphics[width=.13\textwidth]{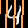} &
        \includegraphics[width=.13\textwidth]{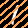} &
        \includegraphics[width=.13\textwidth]{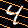} \\
        
        \includegraphics[width=.13\textwidth]{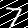} &
        \includegraphics[width=.13\textwidth]{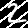} &
        \includegraphics[width=.13\textwidth]{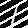} &
        \includegraphics[width=.13\textwidth]{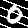} &
        \includegraphics[width=.13\textwidth]{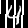} &
        \includegraphics[width=.13\textwidth]{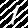} &
        \includegraphics[width=.13\textwidth]{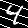} \\

        \includegraphics[width=.13\textwidth]{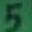} &
        \includegraphics[width=.13\textwidth]{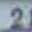} &
        \includegraphics[width=.13\textwidth]{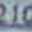} &
        \includegraphics[width=.13\textwidth]{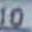} &
        \includegraphics[width=.13\textwidth]{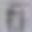} &
        \includegraphics[width=.13\textwidth]{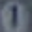} &
        \includegraphics[width=.13\textwidth]{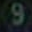} \\
        
        \includegraphics[width=.13\textwidth]{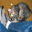} &
        \includegraphics[width=.13\textwidth]{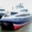} &
        \includegraphics[width=.13\textwidth]{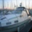} &
        \includegraphics[width=.13\textwidth]{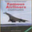} &
        \includegraphics[width=.13\textwidth]{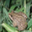} &
        \includegraphics[width=.13\textwidth]{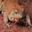} &
        \includegraphics[width=.13\textwidth]{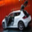} \\

        \includegraphics[width=.13\textwidth]{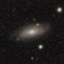} &
        \includegraphics[width=.13\textwidth]{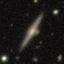} &
        \includegraphics[width=.13\textwidth]{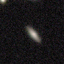} &
        \includegraphics[width=.13\textwidth]{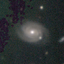} &
        \includegraphics[width=.13\textwidth]{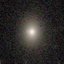} &
        \includegraphics[width=.13\textwidth]{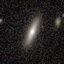} &
        \includegraphics[width=.13\textwidth]{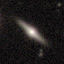} \\
        
    \end{tabular}
\caption{Samples per row from each dataset. From top to bottom: MNIST, MNIST with yellow stripes, MNIST with white stripes, SVHN, CIFAR-10, GalaxyMNIST.}
\label{fig:datasets}
\end{figure}

We conducted experiments on six datasets, including variations of MNIST \cite{lecun1998mnist} and other widely-used image classification datasets. Samples from each dataset are visualized in Figure \ref{fig:datasets}. For all models, the images were resized to 224x224 using bilinear interpolation with the Python Image Library (PIL) \cite{umesh2012image}.

\paragraph{MNIST:}  

The MNIST dataset consists of 60,000 training images and 10,000 test images of handwritten digits, each of size 28x28. These grayscale images were expanded into three-channel images by treating them as L-mode images and converting them to RGB using PIL.

\paragraph{MNIST with Yellow Stripes:}  
This dataset is a modification of the original MNIST, where yellow stripes were added across the top of each digit. These stripes can be seen as a type of augmentation technique or even as a form of adversarial attack, designed to make the images deviate from the original training set. The yellow stripes partially obscure the digits but remain easily recognizable.

\paragraph{MNIST with White Stripes:}  
This MNIST variation includes white stripes across the top of each image, making the digits harder to recognize as the stripes overlap with the digits, often blending in due to the same color. This modification presents a greater challenge than the yellow stripes.

\paragraph{SVHN (Street View House Numbers):}  
The SVHN dataset \cite{netzer2011reading} consists of 73,257 training and 26,032 testing images of real-world house numbers captured from Google Street View. Each 32x32 RGB image contains one or more digits, but the task is to classify the digit in the center, adding complexity compared to simpler datasets like MNIST.

\paragraph{CIFAR-10:}  
The CIFAR-10 dataset \cite{krizhevsky2014cifar} contains 60,000 32x32 color images, with 50,000 for training and 10,000 for testing, across 10 different classes. Each class represents a distinct object category such as airplanes, cars, or animals. 

\paragraph{Galaxy MNIST:}  
GalaxyMNIST \cite{walmsley2022galaxy} contains 10,000 galaxy images (64x64), categorized into four morphological classes: smooth and round, smooth and cigar-shaped, edge-on disk, and unbarred spiral. Derived from Galaxy Zoo DECaLS, it offers a balanced 80/20 train/test split. We use GalaxyMNIST due to its complexity and domain shift, as its astronomical images differ significantly from natural images, allowing us to assess model adaptation in specialized domains.

\paragraph{} These datasets introduce varying levels of complexity and domain shifts compared to the large-scale natural image datasets typically used to pre-train foundation models. While MNIST and CIFAR-10 are relatively simple, the MNIST variations introduce perturbations that challenge the vision foundation models' generalization capabilities. Furthermore, datasets like SVHN and GalaxyMNIST present distinct challenges that highlight the issue of distribution shift. SVHN contains real-world digit images, often including surrounding digits and complex backgrounds, complicating the model's ability to isolate relevant information. GalaxyMNIST, meanwhile, introduces astronomical patterns, which considerably differ from the natural image distributions on which foundation models were trained. This divergence leads to suboptimal performance, making targeted fine-tuning essential, especially when labeled data is limited and unlabeled data is insufficient to train a new foundation model from scratch.

\begin{figure}[ht]
    \centering
    \includegraphics[width=\textwidth]{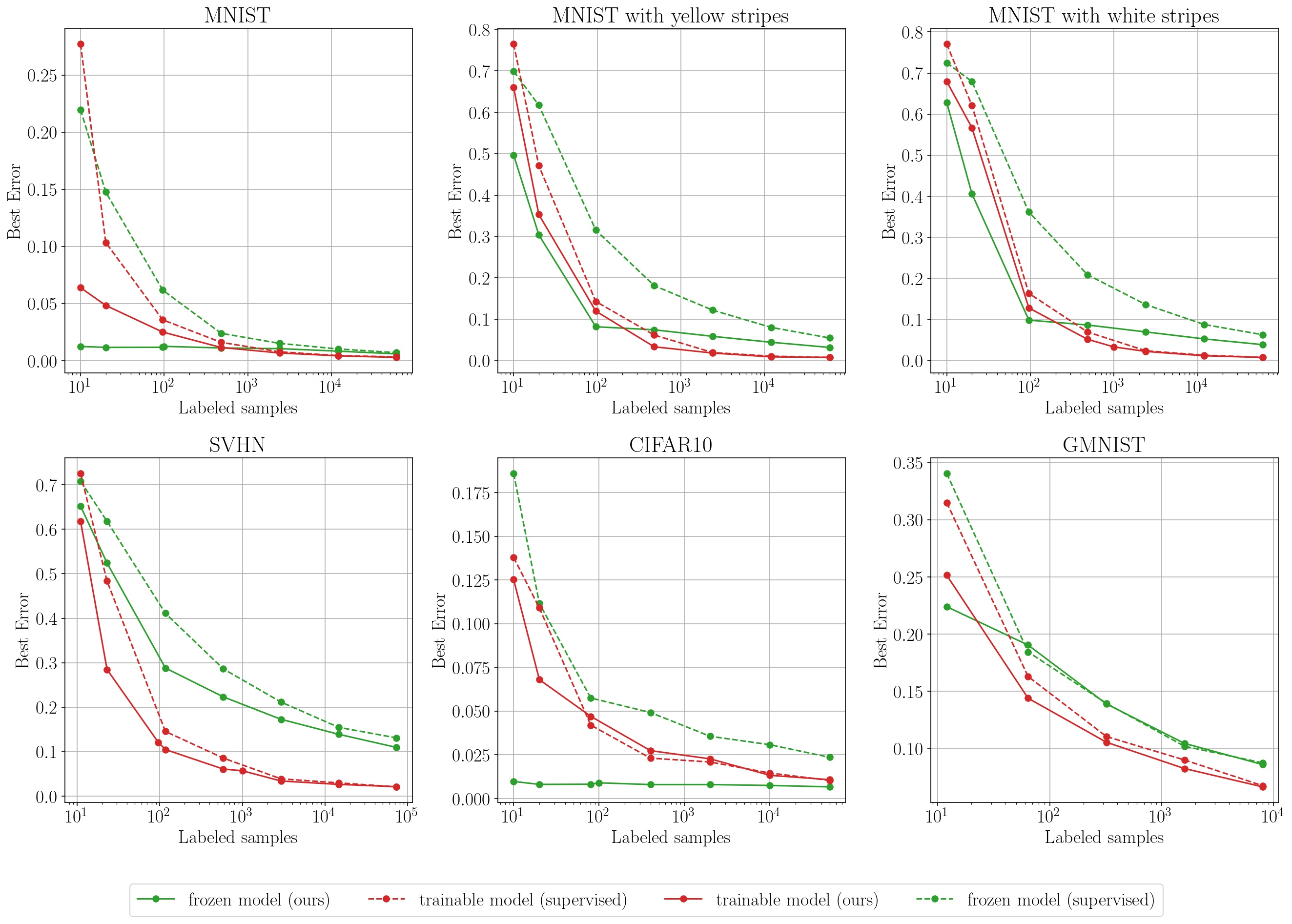}
    \caption{RADIOv2 model results. The y-axis represents the best error rate (1 - accuracy), and the x-axis represents the number of labeled samples. For each dataset, the classifier is trained with both supervised learning and our proposed method for frozen and trainable backbone.}
    \label{fig:RadioV2_results}
\end{figure}

\begin{figure}[ht]
    \centering
    \includegraphics[width=\textwidth]{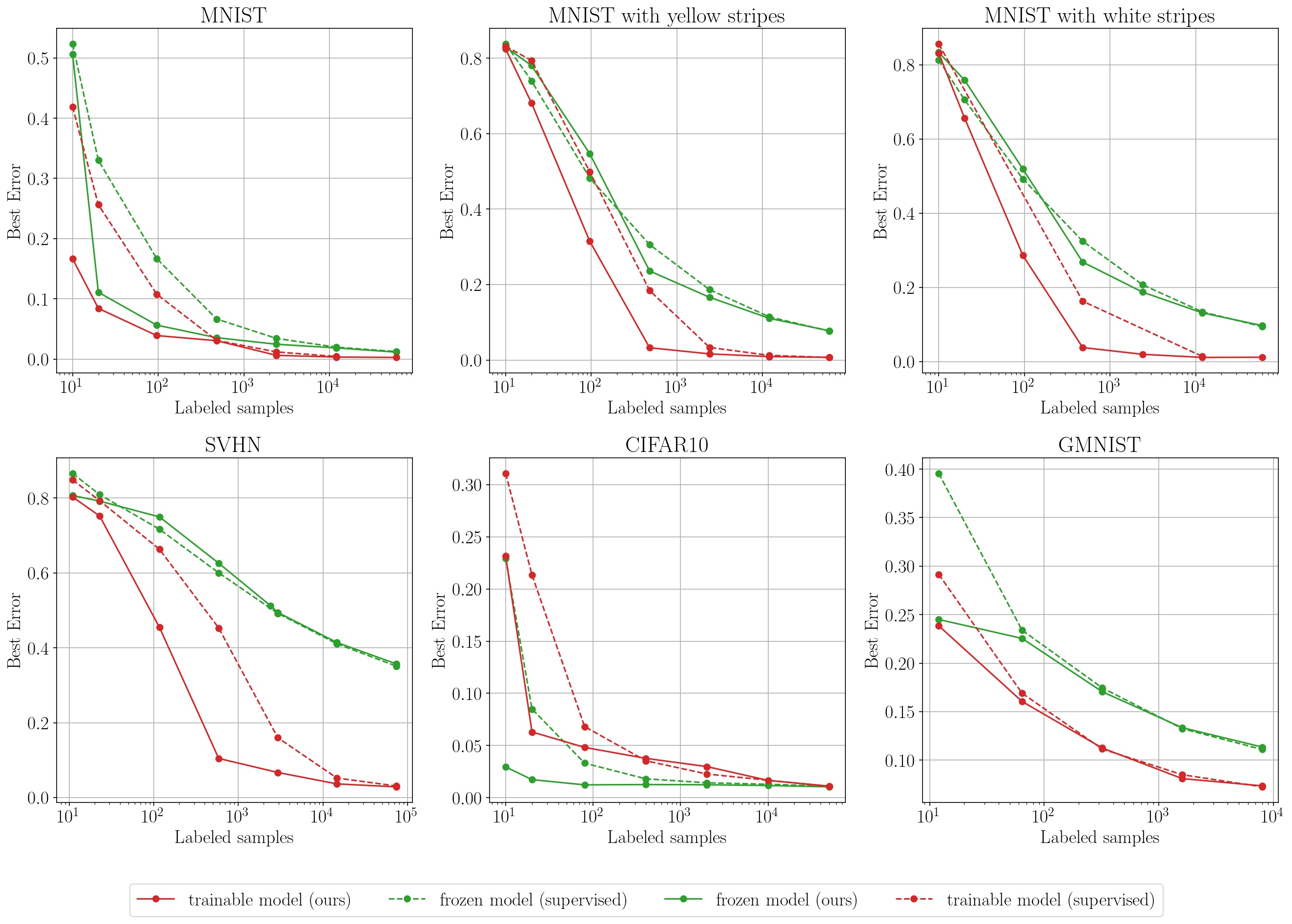}
    \caption{DINOv2 model results. The y-axis represents the best error rate (1 - accuracy), and the x-axis represents the number of labeled samples. For each dataset, the classifier is trained with both supervised learning and our proposed method for frozen and trainable backbone.}
    \label{fig:DinoV2_results}
\end{figure}

\begin{figure}[ht]
    \centering
    \includegraphics[width=\textwidth]{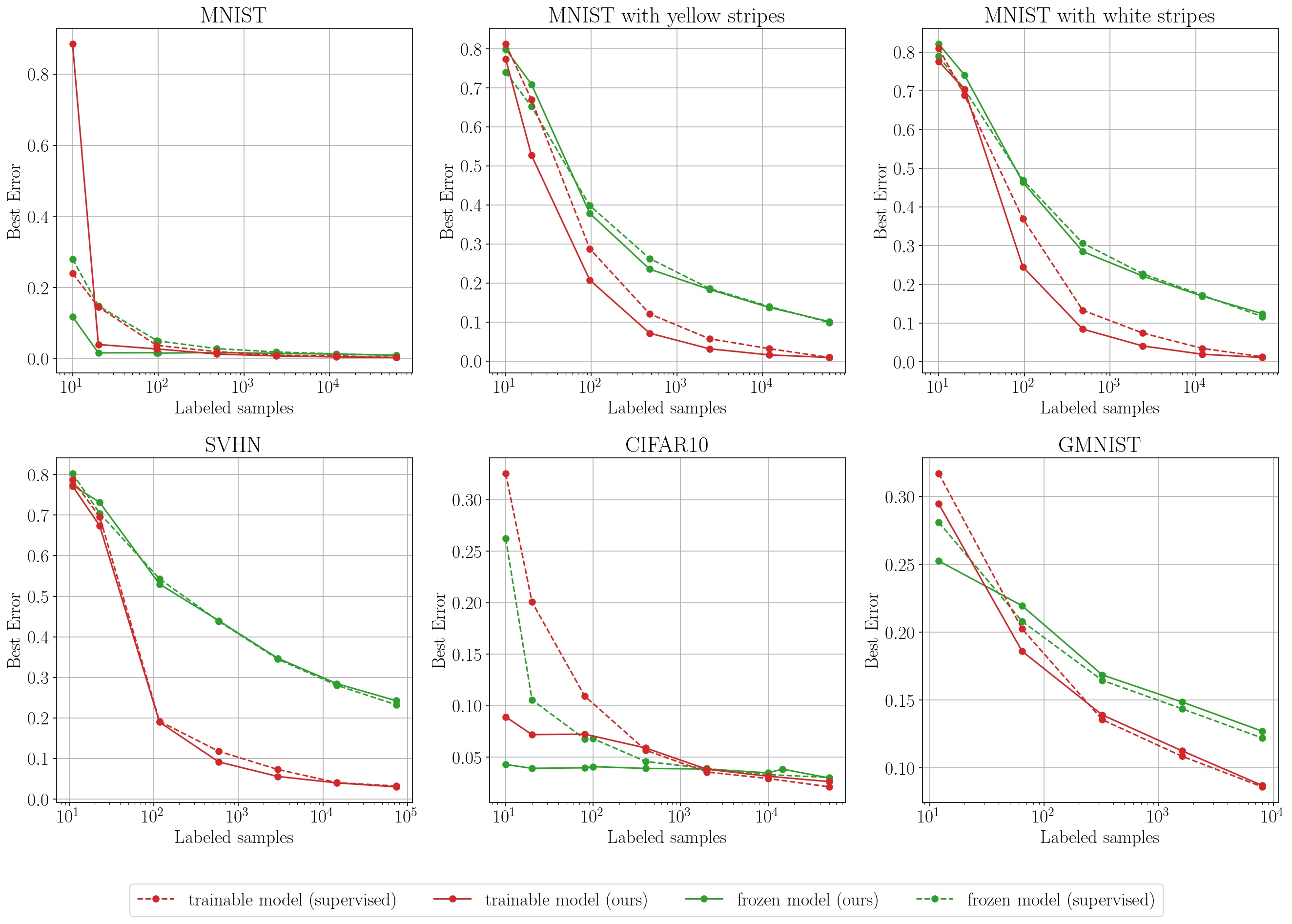}
    \caption{CLIP model results. The y-axis represents the best error rate (1 - accuracy), and the x-axis represents the number of labeled samples. For each dataset, the classifier is trained with both supervised learning and our proposed method for frozen and trainable backbone.}
    \label{fig:Clip_results}
\end{figure}

\section{Results and Discussion}
We evaluated the performance of our proposed semi-supervised fine-tuning method across six datasets: MNIST, MNIST with yellow stripes, MNIST with white stripes, SVHN, CIFAR-10, and GalaxyMNIST. Experiments were conducted with both \textbf{trainable} and \textbf{frozen} pre-trained backbones using three models: RADIOv2.5 Base \cite{Ranzinger_2024_CVPR}, DINOv2 Small \cite{oquab2023Dinov2}, and CLIP Base \cite{ilharco_gabriel_2021_5143773, Radford2021LearningTV, cherti2023reproducible}. We focused on comparing our \textbf{semi-supervised method} to the \textbf{purely supervised fine-tuning} baseline, as it provides a well-established point of reference. Methods like LoRA\cite{hu2021lora} were left out of the scope of this work to avoid introducing additional parameters that would require extensive tuning, allowing us to prioritize a clearer evaluation of our semi-supervised fine-tuning approach. For all baselines in this paper (supervised or our method), we use pre-trained vision foundation models to ensure fair comparisons across methods. Results for each backbone are shown in Figures \ref{fig:RadioV2_results}, \ref{fig:DinoV2_results}, and \ref{fig:Clip_results}.

\subsection{Performance Across Datasets}
Overall, our semi-supervised method consistently improves the performance across all datasets compared to the purely supervised fine-tuning approach. This is particularly evident in the low-labeled data regime (10-100 samples), where our method leverages the combination of labeled and unlabeled data to enhance the learning process. For simpler datasets like MNIST and CIFAR-10, frozen backbones benefit significantly from our method, while for more complex datasets like SVHN and GalaxyMNIST, the trainable models take advantage of the extra flexibility provided by fine-tuning.

\paragraph{MNIST and CIFAR-10: Frozen Models Excel} 
In the MNIST and CIFAR-10 datasets, frozen backbones consistently outperform trainable models under our semi-supervised method, especially in the low-labeled data regime. The simplicity of these datasets, where the content-label relationship is straightforward, allows the frozen models to maintain strong initial representations, and further fine-tuning often leads to overfitting or suboptimal adjustments.

RADIOv2 consistently achieves the best performance across all models, particularly in the frozen setup. As the number of labeled samples increases, trainable models start to close the gap, but frozen RADIOv2 remains competitive. CLIP shows similar patterns, although the performance improvement is less significant compared to RADIOv2; DINOv2, being smaller than RADIOv2 and CLIP, has its frozen backbone underperforming compared to the trainable one under our method. This suggests that RADIOv2’s pre-trained representations are better suited to these datasets.

\paragraph{MNIST with Yellow and White Stripes: Trainable Overtakes for RADIOv2}
For MNIST with yellow and white stripes, we observe a different dynamic. In the low-data regime, frozen models under our method still perform well, particularly for RADIOv2. However, as more labeled data becomes available, the trainable backbone trained with semi-supervision starts to outperform the frozen one for RADIOv2. This indicates that the added complexity introduced by the stripes (which obscure part of the digits) requires the backbone to adapt its features to fully capture the content-label relationship.

For CLIP and DINOv2, trainable backbones perform better consistently, even in the low-data regime, suggesting that these models require more flexibility to handle the added complexity. Again, RADIOv2 shows the strongest overall performance, indicating that its initial representations are more resilient to perturbations like the stripes.

\paragraph{SVHN and GalaxyMNIST: Trainable Models Are Essential}
In more complex datasets like SVHN and GalaxyMNIST, the trainable models clearly outperform the frozen models across all backbones. These datasets exhibit larger intra-class variability and a significant distribution shift from the pre-trained representations. In these cases, fine-tuning the backbone is necessary to adapt the model’s features to the specific task, especially as more labeled samples become available.

Our semi-supervised method shows  improvements in the low-labeled data regime for both frozen and trainable models, though the trainable backbones benefit much more. RADIOv2 continues to show the best performance overall, followed by CLIP and then DINOv2. Our method’s improvements on CLIP are rather limited, which means that the model does not benefit much from the extra unlabeled samples. In contrast, RADIOv2 demonstrates greater capacity to leverage unlabeled data, making it the most adaptable model across datasets.

\subsection{Discussion: Frozen vs Trainable with Our Method}
Our semi-supervised method reveals interesting dynamics between frozen and trainable models. In simpler datasets like MNIST and CIFAR-10, frozen models benefit more from our method, likely because the backbone representations are already well-suited to the task, and fine-tuning may introduce unnecessary changes. This is particularly true for RADIOv2, which consistently outperforms the other models in frozen setups. However, as dataset complexity increases (e.g., MNIST with stripes, SVHN, GalaxyMNIST), fine-tuning backbones become necessary. The trainable models can adapt to more complex content-label relationships and handle the increased intra-class variability, especially in datasets with significant distribution shifts like GalaxyMNIST. 

Overall, our method improves the scores compared to the purely supervised fine-tuning approach for most of the studied datasets with both frozen and trainable backbones.

\section{Conclusion}

This paper emphasizes the critical role of fine-tuning in adapting foundation models to specific downstream tasks, especially in scientific domains with limited labeled data. By leveraging content-style decomposition within an information-theoretic framework, we can effectively tailor the latent representations of foundation models, ensuring their suitability for specific applications. Our findings underscore the importance of aligning model training with the ultimate objectives of downstream tasks to achieve optimal performance.

Our experiments demonstrate that the proposed semi-supervised approach improves performance across both frozen and trainable backbones. The method consistently delivers better results than purely supervised fine-tuning baselines in the majority of the cases. 
Our method offers a step toward mitigating distribution shift, particularly in the early stages of fine-tuning. However, further research is required to fully understand how different backbone architectures respond to domain shifts and how unlabeled data can be leveraged more effectively in various scientific tasks.

Future work will focus on extending this framework to image reconstruction tasks, expanding the scope of downstream applications and studied foundation models, and exploring alternative fine-tuning strategies to address distribution mismatch effectively.

\clearpage
\bibliographystyle{plain} 
\bibliography{references}

\clearpage
\appendix
\section{Information-Theoretic Details}
\label{appendix_A}

\subsection*{Content Decoder Loss}
For the content decoder, the Kullback-Leibler (KL) divergence measures the difference between the true distribution \( p_{c_{a_x}} ({\bf c}_{a_x}) \) of the content label and the model's predicted distribution \( p_{\theta_{c_{a_x}}}(\mathbf{c}_{a_x}) \). The KL divergence is defined as:
\begin{equation}
    \mathcal{D}_{c_{a_x}}: = \mathbb{D}_{KL}(p_{c_{a_x}}(\mathbf{c}_{a_x})
||
p_{\theta_{c_{a_x}}}(\mathbf{c}_{a_x} )) = \mathbb{E}_{p_{c_{a_x}}(\mathbf{c}_{a_x})} \left[ \log \frac{p_{c_{a_x}}(\mathbf{c}_{a_x})}{p_{\theta_{c_{a_x}}}(\mathbf{c}_{a_x} )} \right].
\end{equation}
This term ensures that the model’s content prediction aligns with the true content distribution.

Additionally, we include the cross-entropy loss for the supervised content prediction:
\begin{equation}
    \mathcal{L}_{\text{CE}} = -\mathbb{E}_{p(\mathbf{c}_{a_x}, \mathbf{x})} \left[ \log p_{\theta_{c_{a_x}}}(\mathbf{c}_{a_x}) \right].
\end{equation}

\subsection*{Style Decoder Loss}
For the style decoder, the KL divergence measures the difference between the true prior distribution \( p_{s_{a_x}}(\mathbf{s}_{a_x})\) and the model’s predicted distribution \( p_{\theta_{s_{a_x}}}(\mathbf{s}_{a_x})\). The KL divergence is defined as:
\begin{equation}
    \mathcal{D}_{s_{a_x}}: =\mathbb{D}_{KL}(p_{s_{a_x}}(\mathbf{s}_{a_x})
||
p_{\theta_{s_{a_x}}}(\mathbf{s}_{a_x} )) = \mathbb{E}_{p_{s_{a_x}}(\mathbf{s}_{a_x})} \left[ \log \frac{p_{s_{a_x}}(\mathbf{s}_{a_x})}{p_{\theta_{s_{a_x}}}(\mathbf{s}_{a_x})} \right].
\end{equation}

\subsection*{\texttt{[CLS]} Token Decoder Loss}
For the \texttt{[CLS]} token, the KL divergence is computed between the true prior distribution \( p_{y_x} (\mathbf{y}_{x})\) and the model’s predicted distribution \( p_{\theta_{y_x}}(\mathbf{y}_x) \):
\begin{equation}
    \mathcal{D}_{y_x}: = \mathbb{D}_{KL}(p_{y_{x}}(\mathbf{y}_{x})
||
p_{\theta_{y_{x}}}(\mathbf{y}_{x} )) = \mathbb{E}_{p_{y_x}(\mathbf{y}_x)} \left[ \log \frac{p_{y_x} (\mathbf{y}_x)}{p_{\theta_{y_x}} (\mathbf{y}_x)} \right].
\end{equation}

Additionally, we include the cosine similarity loss for the reconstruction of the \texttt{[CLS]} token:
\begin{equation}
    \mathcal{L}_{y_x} \left( \tilde{\mathbf{y}}_x, \hat{\mathbf{y}}_x \right) = 1 - \cos \left( \tilde{\mathbf{y}}_x, \hat{\mathbf{y}}_x \right).
\end{equation}
This loss can be considered as a conditional cross-entropy for the reconstrcution. 

\subsection*{Total Loss Function}
The total loss combines the content, style, and \texttt{[CLS]} token losses, along with their respective KL divergence regularizers and the cross-entropy loss for supervised content prediction:
\begin{equation}
    \mathcal{L}_{\text{total}} = \mathcal{L}_{\text{CE}} + \lambda_c \mathcal{D}_{c_{a_x}} + \lambda_s \mathcal{D}_{s_{a_x}} + \lambda_y \mathcal{D}_{y_x} + \lambda_{y\hat{y}} \mathcal{L}_{y_x}.
\end{equation}

In our experiments $\lambda_c = \lambda_s =\lambda_y = \lambda_{y\hat{y}} = 1$ following ablation studies conducted in \cite{voloshynovskiy2020variational}.
\section{Model Descriptions}
\label{appendix_B}

This appendix provides a detailed description of the models used in our experiments. The architecture includes an encoder content and style outputs (\(p_{\theta_{{c}_{a_x}}}\left(\mathbf{c}_{a_x} | \mathbf{y}_x\right)\) and \(p_{\theta_{{s}_{a_x}}}\left(\mathbf{s}_{a_x} | \mathbf{y}_x\right)\)), a decoder for reconstruction (\(p_{\theta_{y_x}}(\mathbf{y}_x|\mathbf{c}_{a_x}, \mathbf{s}_{a_x})\)), and three discriminators: one for content (\(\mathcal{D}_{c_{a_x}}\)), one for style (\(\mathcal{D}_{s_{a_x}}\)), and one for the \texttt{CLS} token (\(\mathcal{D}_{y_x}\)). Each component plays a crucial role in ensuring that the model performs well under semi-supervised learning.

\subsection*{Encoder}

The encoder is responsible for processing the input features through a shared MLP structure that maps them into meaningful representations. The shared MLP is shown in Table \ref{tab:encoder}.

\begin{table}[ht]
\centering
\caption{Shared Encoder Structure}
\label{tab:encoder}
\begin{tabular}{cc}
\hline
\textbf{Size} & \textbf{Layer} \\
\hline
\texttt{CLS} Token size & Input \\
8000 & Linear \\
8000 & Leaky ReLU (slope = 0.01) \\
8000 & Dropout (probability = 0.3) \\
\hline
\end{tabular}
\end{table}

\textbf{Content and Style Heads:} After passing through the shared MLP, the features are split into two separate heads: for content \(\mathbf{c}_{a_x}\) and one for style \(\mathbf{s}_{a_x}\) .

\subsubsection*{Content Head}
The content encoder \(p_{\theta_{{c}_{a_x}}}\left(\mathbf{c}_{a_x} | \mathbf{y}_x\right)\) outputs the content representation. The structure of the head is shown in Table \ref{tab:content}.

\begin{table}[ht]
\centering
\caption{Content Head Structure}
\label{tab:content}
\begin{tabular}{cc}
\hline
\textbf{Size} & \textbf{Layer} \\
\hline
8000 & Linear \\
1024 & Leaky ReLU (slope = 0.01) \\
Number of classes & Output (Linear) \\
\hline
\end{tabular}
\end{table}

\subsubsection*{Style Head}
The style encoder \(p_{\theta_{{s}_{a_x}}}\left( \mathbf{s}_{a_x} | \mathbf{y}_x \right)\) generates the style representation. Its head structure is shown in Table \ref{tab:style}.

\begin{table}[ht]
\centering
\caption{Style Head Structure}
\label{tab:style}
\begin{tabular}{cc}
\hline
\textbf{Size} & \textbf{Layer} \\
\hline
8000 & Linear \\
100 & Output \\
\hline
\end{tabular}
\end{table}

\subsection*{Decoder}

The decoder, denoted by \(p_{\theta_{y_x}}(\mathbf{y}_x|\mathbf{c}_{a_x}, \mathbf{s}_{a_x})\), reconstructs the \texttt{CLS} token by stacking the content and style. The decoder structure is shown in Table \ref{tab:decoder}.

\begin{table}[ht]
\centering
\caption{Decoder Structure}
\label{tab:decoder}
\begin{tabular}{cc}
\hline
\textbf{Size} & \textbf{Layer} \\
\hline
Content size + Style size & Input \\
2560 & Linear \\
2560 & Leaky ReLU (slope = 0.01) \\
2560 & Dropout (probability = 0.3) \\
2560 & Linear \\
2560 & Leaky ReLU (slope = 0.01) \\
2560 & Dropout (probability = 0.3) \\
\texttt{CLS} Token size & Output (Linear) \\
\hline
\end{tabular}
\end{table}

\subsection*{Discriminators}

Our method uses binary cross entropy for discriminator losses. The model has three discriminators : one for content \(\mathbf{c}_{a_x}\), one for style \(\mathbf{s}_{a_x}\), and one for the \texttt{CLS} token \(\mathbf{y}_x\). The structure of each discriminator is outlined below.

\subsubsection*{Content Discriminator}
The content discriminator \(\mathcal{D}_{\mathbf{c}_{a_x}}\) ensures that the content vector has one-hot representation for all unpaired inputs. The structure of this discriminator is shown in Table \ref{tab:content_discriminator}.

\begin{table}[ht]
\centering
\caption{Content Discriminator Structure}
\label{tab:content_discriminator}.
\begin{tabular}{cc}
\hline
\textbf{Size} & \textbf{Layer} \\
\hline
Number of classes & Input \\
500 & Linear \\
500 & Linear \\
1 & Sigmoid \\
\hline
\end{tabular}
\end{table}

\subsubsection*{Style Discriminator}
The style discriminator \(\mathcal{D}_{\mathbf{s}_{a_x}}\) ensures that the style representation is Gaussian. The structure of this discriminator is shown in Table \ref{tab:style_discriminator}.

\begin{table}[ht]
\centering
\caption{Style Discriminator Structure}
\label{tab:style_discriminator}
\begin{tabular}{cc}
\hline
\textbf{Size} & \textbf{Layer} \\
\hline
100 & Input \\
50 & Linear \\
500 & Linear \\
1 & Sigmoid \\
\hline
\end{tabular}
\end{table}

\subsubsection*{Class Token Discriminator }
The \texttt{CLS} token discriminator \(\mathcal{D}_{\mathbf{y}_{a_x}}\) evaluates the final \texttt{CLS} representation, ensuring that given the combination of one-hot content and Gaussian style the decoder can generate the \texttt{CLS} token. The structure of this discriminator is shown in Table \ref{tab:classtoken_discriminator}.

\begin{table}[ht]
\centering
\caption{Class Token Discriminator Structure}
\label{tab:classtoken_discriminator}
\begin{tabular}{cc}
\hline
\textbf{Size} & \textbf{Layer} \\
\hline
\texttt{CLS} Token size & Input \\
128 & Linear \\
128 & Leaky ReLU (slope = 0.02) \\
64 & Linear \\
64 & Leaky ReLU (slope = 0.02) \\
32 & Linear \\
32 & Leaky ReLU (slope = 0.02) \\
1 & Sigmoid \\
\hline
\end{tabular}
\end{table}

\subsection*{Training Setup}

The training process involves optimizing the network using the AdamW optimizer and a learning rate of $5 \times 10^{-5}$. A warmup schedule of 0 for components and 0.1 for transformers is applied. Batch sizes are set to 512 for frozen models (unsupervised losses) and 32 for trainable models. 

The following losses are used:
\begin{itemize}
    \item \textbf{Adversarial losses:} Applied for all discriminators.
    \item \textbf{Supervised loss:} Cross-entropy loss applied to the content head.
    \item \textbf{Reconstruction loss:} Cosine similarity loss for the \texttt{CLS} token reconstruction.
\end{itemize}

\section{t-SNE in the classifier features for RADIOv2}
\label{appendix_C}

Here, we visualize classifier features before the last linear layer, where the features have a dimensionality of 1024, for both frozen and trainable RADIOv2 models. We first apply PCA to reduce the features to 5 components, followed by t-SNE for visualization. The images are from the test set, and the colors correspond to the classes. The classifier is fully trained. The upper row shows our model, and the lower row shows the supervised method. The images are ordered by an increasing number of labeled data available for supervised updates.

\begin{figure}
    \centering
    \renewcommand{\arraystretch}{1} %
    \begin{tikzpicture}
         
        \node (image) at (0,0) {\includegraphics[width=0.8\textwidth,trim=0 0 200 50,clip]{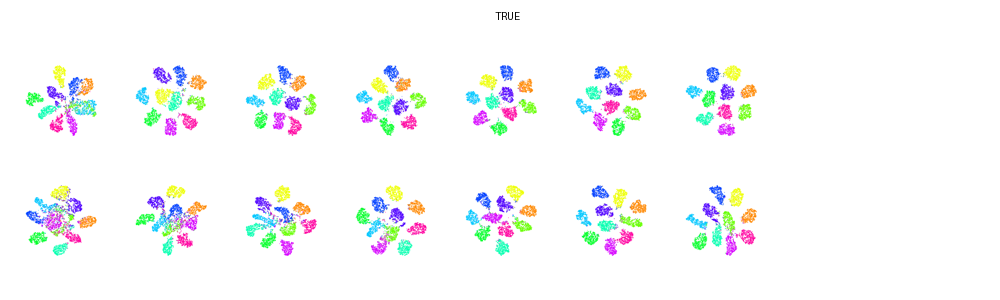}};

        \draw[->, thick] (-5,-1.8) -- (5,-1.8) node[midway, below] {increasing available labeled data};

        \node at (-7,-0.5) {\textbf{supervised}};

        \node at (-7,1) {\textbf{our method}};
    \end{tikzpicture}
\caption{Latent space (dimensionality of 1024) of the classifier for the MNIST dataset with a \textbf{frozen} backbone. The first row shows our method, and the second row shows the supervised method. From left to right, the total number of labeled samples increases: 10, 19, 96, 480, 2,400, 12,000, 60,000.}
    \label{fig:tsne_cosine_true}
\end{figure}
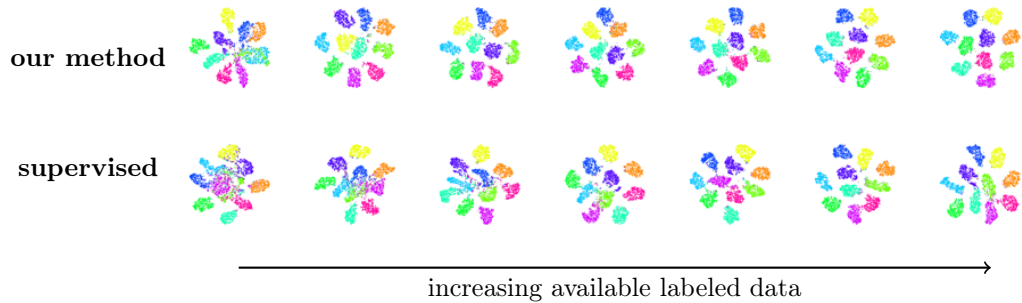

\begin{figure}
    \centering
    \renewcommand{\arraystretch}{1} %
    \begin{tikzpicture}
         
        \node (image) at (0,0) {\includegraphics[width=0.8\textwidth,trim=0 0 200 50,clip]{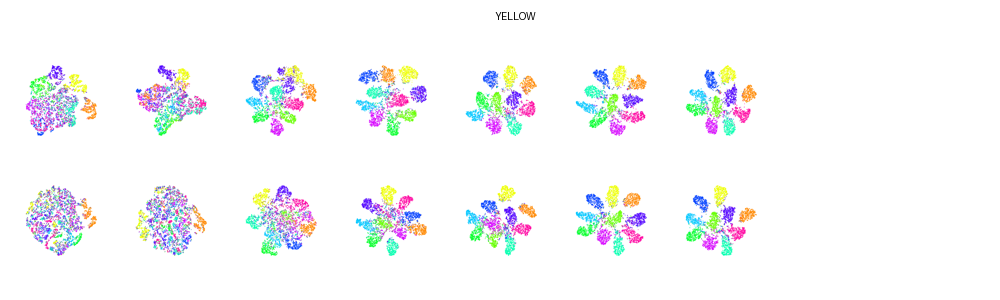}};

        \draw[->, thick] (-5,-1.8) -- (5,-1.8) node[midway, below] {increasing available labeled data};

        \node at (-7,-0.5) {\textbf{supervised}};

        \node at (-7,1) {\textbf{our method}};
    \end{tikzpicture}
\caption{Latent space (dimensionality of 1024) of the classifier for the MNIST dataset with yellow stripes with a \textbf{frozen} backbone. The first row shows our method, and the second row shows the supervised method. From left to right, the total number of labeled samples used for training increases: 10, 19, 96, 480, 2,400, 12,000, 60,000.}
    \label{fig:tsne_cosine_yellow}
\end{figure}

\begin{figure}
    \centering
    \renewcommand{\arraystretch}{1} %
    \begin{tikzpicture}
         
        \node (image) at (0,0) {\includegraphics[width=0.8\textwidth,trim=0 0 200 50,clip]{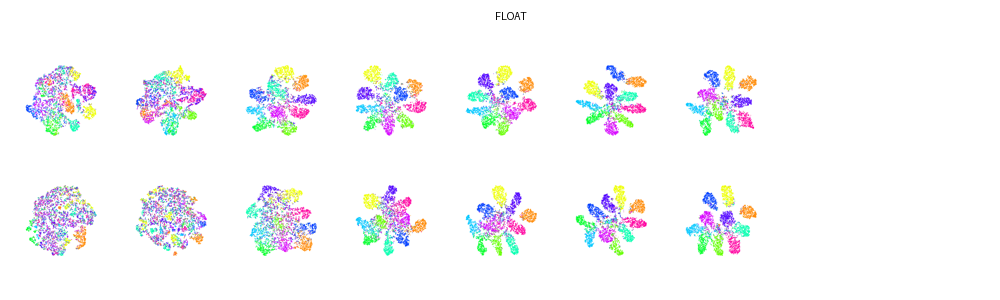}};

        \draw[->, thick] (-5,-1.8) -- (5,-1.8) node[midway, below] {increasing available labeled data};

        \node at (-7,-0.5) {\textbf{supervised}};

        \node at (-7,1) {\textbf{our method}};
    \end{tikzpicture}
\caption{Latent space (dimensionality of 1024) of the classifier for the MNIST dataset with white stripes with a \textbf{frozen} backbone. The first row shows our method, and the second row shows the supervised method. From left to right, the total number of labeled samples used for training increases: 10, 19, 96, 480, 2,400, 12,000, 60,000.}
    \label{fig:tsne_cosine_float}
\end{figure}

\begin{figure}
    \centering
    \renewcommand{\arraystretch}{1} %
    \begin{tikzpicture}
         
        \node (image) at (0,0) {\includegraphics[width=0.8\textwidth,trim=0 0 200 50,clip]{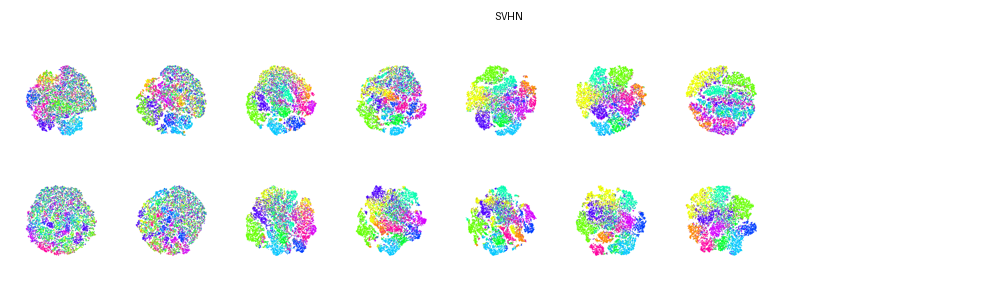}};

        \draw[->, thick] (-5,-1.8) -- (5,-1.8) node[midway, below] {increasing available labeled data};

        \node at (-7,-0.5) {\textbf{supervised}};

        \node at (-7,1) {\textbf{our method}};
    \end{tikzpicture}
    
\caption{Latent space (dimensionality of 1024) of the classifier for the SVHN dataset with a \textbf{frozen} backbone. The first row shows our method, and the second row shows the supervised method. From left to right, the total number of labeled samples used for training increases: 10, 23, 117, 586, 2,930, 14,651, 73,257.}
    \label{fig:tsne_cosine_SVHN}
\end{figure}

\begin{figure}
    \centering
    \renewcommand{\arraystretch}{1} %
    \begin{tikzpicture}
         
        \node (image) at (0,0) {\includegraphics[width=0.8\textwidth,trim=0 0 200 50,clip]{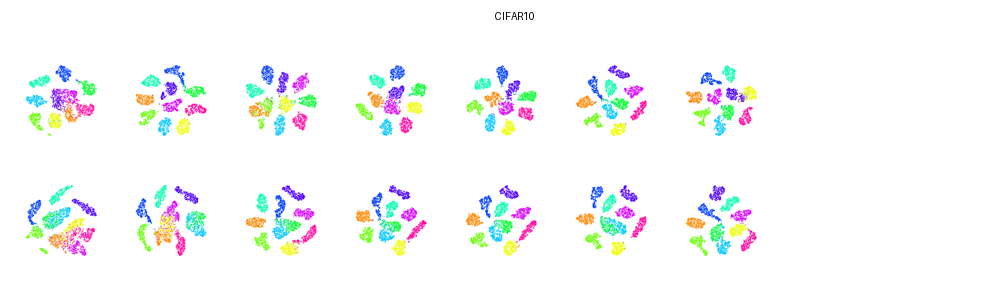}};

        \draw[->, thick] (-5,-1.8) -- (5,-1.8) node[midway, below] {increasing available labeled data};

        \node at (-7,-0.5) {\textbf{supervised}};

        \node at (-7,1) {\textbf{our method}};
    \end{tikzpicture}
    
\caption{Latent space (dimensionality of 1024) of the classifier for the CIFAR-10 dataset with a \textbf{frozen} backbone. The first row shows our method, and the second row shows the supervised method. From left to right, the total number of labeled samples used for training increases: 10, 16, 80, 400, 2,000, 10,000, 50,000.}
    \label{fig:tsne_cosine_cifar10}
\end{figure}

\begin{figure}
    \centering
    \renewcommand{\arraystretch}{1} %
    \begin{tikzpicture}
         
        \node (image) at (0,0) {\includegraphics[width=0.8\textwidth,trim=0 0 450 50,clip]{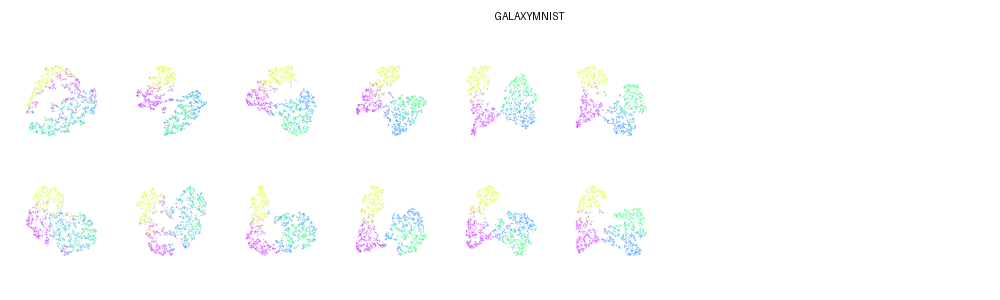}};

        \draw[->, thick] (-5,-2.0) -- (5,-2.0) node[midway, below] {increasing available labeled data};

        \node at (-7,-0.5) {\textbf{supervised}};

        \node at (-7,1) {\textbf{our method}};
    \end{tikzpicture}
    
\caption{Latent space (dimensionality of 1024) of the classifier for the GalaxyMNIST dataset with a \textbf{frozen} backbone. The first row shows our method, and the second row shows the supervised method. From left to right, the total number of labeled samples used for training increases: 4, 12, 64, 320, 1,600}
    \label{fig:tsne_cosine_galaxymnist}
\end{figure}

\begin{figure}
    \centering
    \renewcommand{\arraystretch}{1} %
    \begin{tikzpicture}
         
        \node (image) at (0,0) {\includegraphics[width=0.8\textwidth,trim=0 0 200 50,clip]{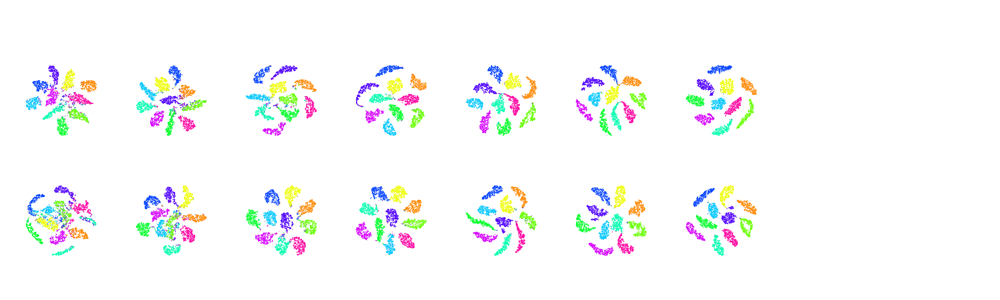}};

        \draw[->, thick] (-5,-1.8) -- (5,-1.8) node[midway, below] {increasing available labeled data};

        \node at (-7,-0.5) {\textbf{supervised}};

        \node at (-7,1) {\textbf{our method}};
    \end{tikzpicture}
    
\caption{Latent space (dimensionality of 1024) of the classifier for the MNIST dataset with a \textbf{trainable} backbone. The first row shows our method, and the second row shows the supervised method. From left to right, the total number of labeled samples increases: 10, 19, 96, 480, 2,400, 12,000, 60,000.}
    \label{fig:tsne_trainable_true}
\end{figure}

\begin{figure}
    \centering
    \renewcommand{\arraystretch}{1} %
    \begin{tikzpicture}
         
        \node (image) at (0,0) {\includegraphics[width=0.8\textwidth,trim=0 0 200 50,clip]{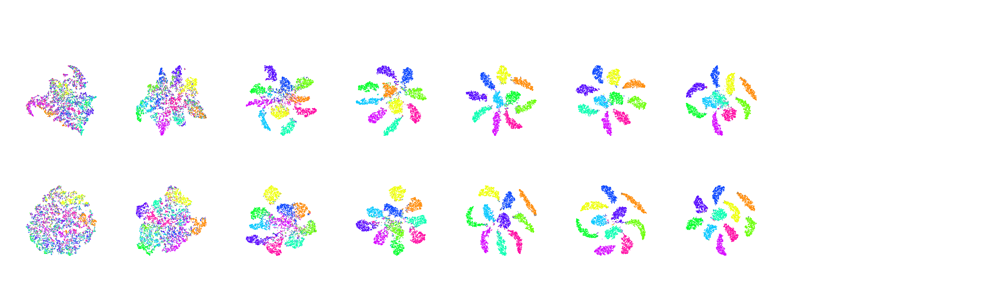}};

        \draw[->, thick] (-5,-1.8) -- (5,-1.8) node[midway, below] {increasing available labeled data};

        \node at (-7,-0.5) {\textbf{supervised}};

        \node at (-7,1) {\textbf{our method}};
    \end{tikzpicture}
    
\caption{Latent space (dimensionality of 1024) of the classifier for the MNIST dataset with yellow stripes with a \textbf{trainable} backbone. The first row shows our method, and the second row shows the supervised method. From left to right, the total number of labeled samples used for training increases: 10, 19, 96, 480, 2,400, 12,000, 60,000.}
    \label{fig:tsne_trainable_yellow}
\end{figure}

\begin{figure}
    \centering
    \renewcommand{\arraystretch}{1} %
    \begin{tikzpicture}
         
        \node (image) at (0,0) {\includegraphics[width=0.8\textwidth,trim=0 0 200 50,clip]{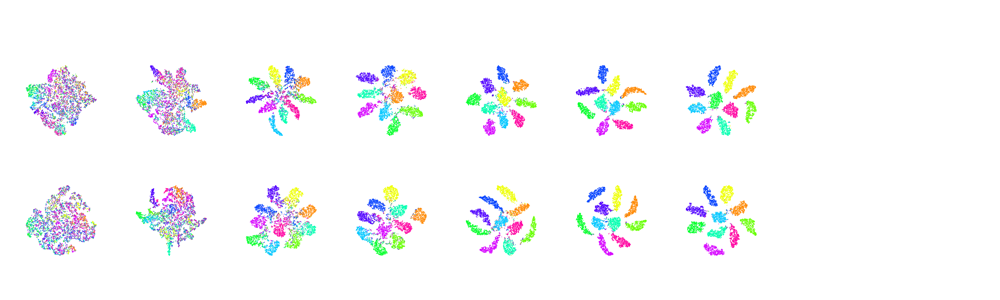}};

        \draw[->, thick] (-5,-1.8) -- (5,-1.8) node[midway, below] {increasing available labeled data};

        \node at (-7,-0.5) {\textbf{our method}};

        \node at (-7,1) {\textbf{our method}};
    \end{tikzpicture}
    
\caption{Latent space (dimensionality of 1024) of the classifier for the MNIST dataset with white stripes with a \textbf{trainable} backbone. The first row shows our method, and the second row shows the supervised method. From left to right, the total number of labeled samples used for training increases: 10, 19, 96, 480, 2,400, 12,000, 60,000.}
    \label{fig:tsne_trainable_white}
\end{figure}

\begin{figure}
    \centering
    \renewcommand{\arraystretch}{1} %
    \begin{tikzpicture}
         
        \node (image) at (0,0) {\includegraphics[width=0.8\textwidth,trim=0 0 200 50,clip]{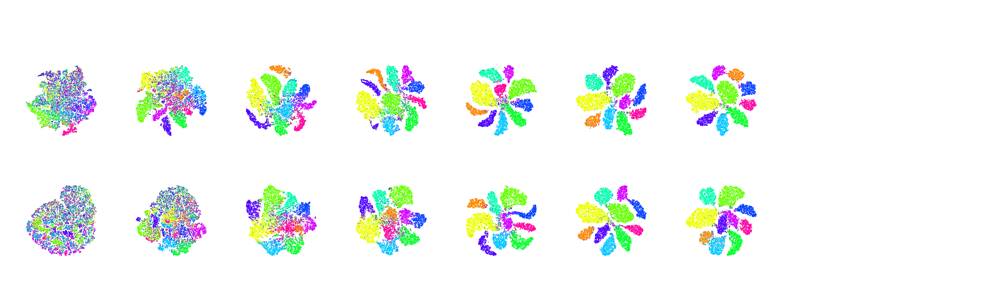}};

        \draw[->, thick] (-5,-1.8) -- (5,-1.8) node[midway, below] {increasing available labeled data};

        \node at (-7,-0.5) {\textbf{supervised}};

        \node at (-7,1) {\textbf{our method}};
    \end{tikzpicture}
    
\caption{Latent space (dimensionality of 1024) of the classifier for the SVHN dataset with a \textbf{trainable} backbone. The first row shows our method, and the second row shows the supervised method. From left to right, the total number of labeled samples used for training increases: 10, 23, 117, 586, 2,930, 14,651, 73,257.}
    \label{fig:tsne_trainable_svhn}
\end{figure}

\begin{figure}
    \centering
    \renewcommand{\arraystretch}{1} %
    \begin{tikzpicture}
         
        \node (image) at (0,0) {\includegraphics[width=0.8\textwidth,trim=0 0 200 50,clip]{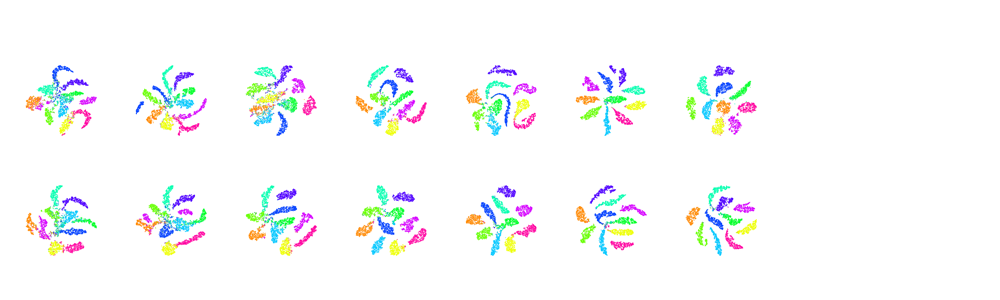}};

        \draw[->, thick] (-5,-1.8) -- (5,-1.8) node[midway, below] {increasing available labeled data};

        \node at (-7,-0.5) {\textbf{supervised}};

        \node at (-7,1) {\textbf{our method}};
    \end{tikzpicture}
    
\caption{Latent space (dimensionality of 1024) of the classifier for the CIFAR-10 dataset with a \textbf{trainable} backbone. The first row shows our method, and the second row shows the supervised method. From left to right, the total number of labeled samples used for training increases: 10, 16, 80, 400, 2,000, 10,000, 50,000.}
    \label{fig:tsne_trainable_cifar10}
\end{figure}

\begin{figure}
    \centering
    \renewcommand{\arraystretch}{1} %
    \begin{tikzpicture}
         
        \node (image) at (0,0) {\includegraphics[width=0.8\textwidth,trim=0 0 450 50,clip]{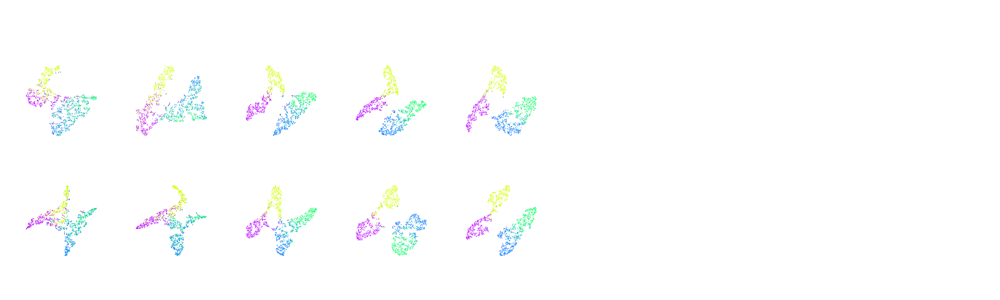}};

        \draw[->, thick] (-5,-2.0) -- (5,-2.0) node[midway, below] {increasing available labeled data};

        \node at (-7,-0.5) {\textbf{supervised}};

        \node at (-7,1) {\textbf{our method}};
    \end{tikzpicture}
    
    \caption{Latent space (dimensionality of 1024) of the classifier for the GalaxyMNIST dataset with a \textbf{trainable} backbone. The first row shows our method, and the second row shows the supervised method. From left to right, the total number of labeled samples used for training increases: 4, 12, 64, 320, 1,600}
    \label{fig:tsne_trainable_galaxymnist}
\end{figure}
\end{document}